\documentclass[letterpaper, 10 pt, conference]{ieeeconf}  % Comment this line out
\IEEEoverridecommandlockouts                              % This command is only
\usepackage{booktabs}% For formal tables
\usepackage{todonotes}
\usepackage{subfigure}
\usepackage[noadjust]{cite}
\usepackage[hidelinks]{hyperref}
\usepackage{caption}
\usepackage{siunitx}
\title{\LARGE \bf The Supernumerary Robotic 3\textsuperscript{rd} Thumb for Skilled Music Tasks}

\author{James Cunningham, Anita Hapsari, Pierre Guilleminot, Ali Shafti, and A. Aldo Faisal
\thanks{Research supported by eNHANCE (\href{http://www.enhance-motion.eu}{http://www.enhance-motion.eu}) under the European Union's Horizon 2020 research and innovation programme grant agreement No 644000.}
\thanks{J. Cunningham, A. Hapsari, P. Guilleminot, A. Shafti and A. A. Faisal are with the Brain and Behaviour Lab, Dept. of Computing and Dept. of Bioengineering, Imperial College London, SW7 2AZ, London, UK.{\tt\small  a.faisal@imperial.ac.uk}}
}

\begin{document}

\maketitle
\thispagestyle{empty}
\pagestyle{empty}

%%%%%%%%%%%%%%%%%%%%%%%%%%%%%%%%%%%%%%%%%%%%%%%%%%%%%%%%%%%%%%%%%%%%%%%%%%%%%%%%
\begin{abstract}

Wearable robotics bring the opportunity to augment human capability and performance, be it through prosthetics, exoskeletons, or supernumerary robotic limbs. The latter concept allows enhancing human performance and assisting them in daily tasks. An important research question is, however, whether the use of such devices can lead to their eventual cognitive embodiment, allowing the user to adapt to them and use them seamlessly as any other limb of their own. This paper describes the creation of a platform to investigate this. Our supernumerary robotic 3\textsuperscript{rd} thumb was created to augment piano playing, allowing a pianist to press piano keys beyond their natural hand-span; thus leading to functional augmentation of their skills and the technical feasibility to play with 11 fingers. The robotic finger employs sensors, motors, and a human interfacing  algorithm to control its movement in real-time. A proof of concept validation experiment has been conducted to show the effectiveness of the robotic finger in playing musical pieces on a grand piano, showing that naive users were able to use it for 11 finger play within a few hours.
\end{abstract}

%%%%%%%%%%%%%%%%%%%%%%%%%%%%%%%%%%%%%%%%%%%%%%%%%%%%%%%%%%%%%%%%%%%%%%%%%%%%%%%%
\section{Introduction}
Human augmentation can be achieved through wearable robotics, these devices enhance human capability by replacing missing limbs \cite{Rouse2013ClutchableConsumption,xiloyannis2017gaussian} giving more power to existing human body parts in the form of exoskeletons \cite{Gopura2009SUEFUL-7:Control,Dollar2008LowerState-of-the-art,Veneman2007DesignRehabilitation} or by providing additional   limbs \cite{Hussain2015DesignExtra-finger,Llorens-Bonilla2014APredictions,Prattichizzo2014TheCapabilities,Parietti2014BracingReduction,Wu2014Bio-artificialFingers,Llorens-Bonilla2012Demonstration-basedLimbs}. This latter category of wearable robotics is referred to as supernumerary robotic limbs (SRL). These are devices that attach to the human body, giving it more degrees of freedom and thus more capabilities. An additional robotic limbs can lead to unprecedented opportunities in augmenting human capability and performance, including reduction of the number of workers required to do a particular task \cite{Llorens-Bonilla2014APredictions,Parietti2014BracingReduction,Llorens-Bonilla2012Demonstration-basedLimbs}, or improvement of the manipulation capabilities of a single hand in grasping an object \cite{Wu2014Bio-artificialFingers,Hussain2015UsingPatients}. Currently most of the research on additional robotic limbs for healthy humans are focused on grasping of objects \cite{Hussain2015DesignExtra-finger,Wu2014Bio-artificialFingers,Hussain2015UsingPatients} and reducing the workload on industrial workers \cite{Llorens-Bonilla2014APredictions,Parietti2014BracingReduction}.

This paper presents efforts to expand the focus of supernumerary robotic limbs to the realm of music technology and more functional tasks within the augmentation of the hand. The goal is thus to develop a supernumerary robotic 3\textsuperscript{rd} thumb (SR3T) finger that can augment musicians in playing piano, including expanding their reach to larger chords, playing keys located beyond the natural reach of a single hand-span, and creating unprecedented complex music that is impossible to play with 10 fingers.
\begin{figure}[tp]
\centering
\includegraphics[width=0.8\columnwidth]{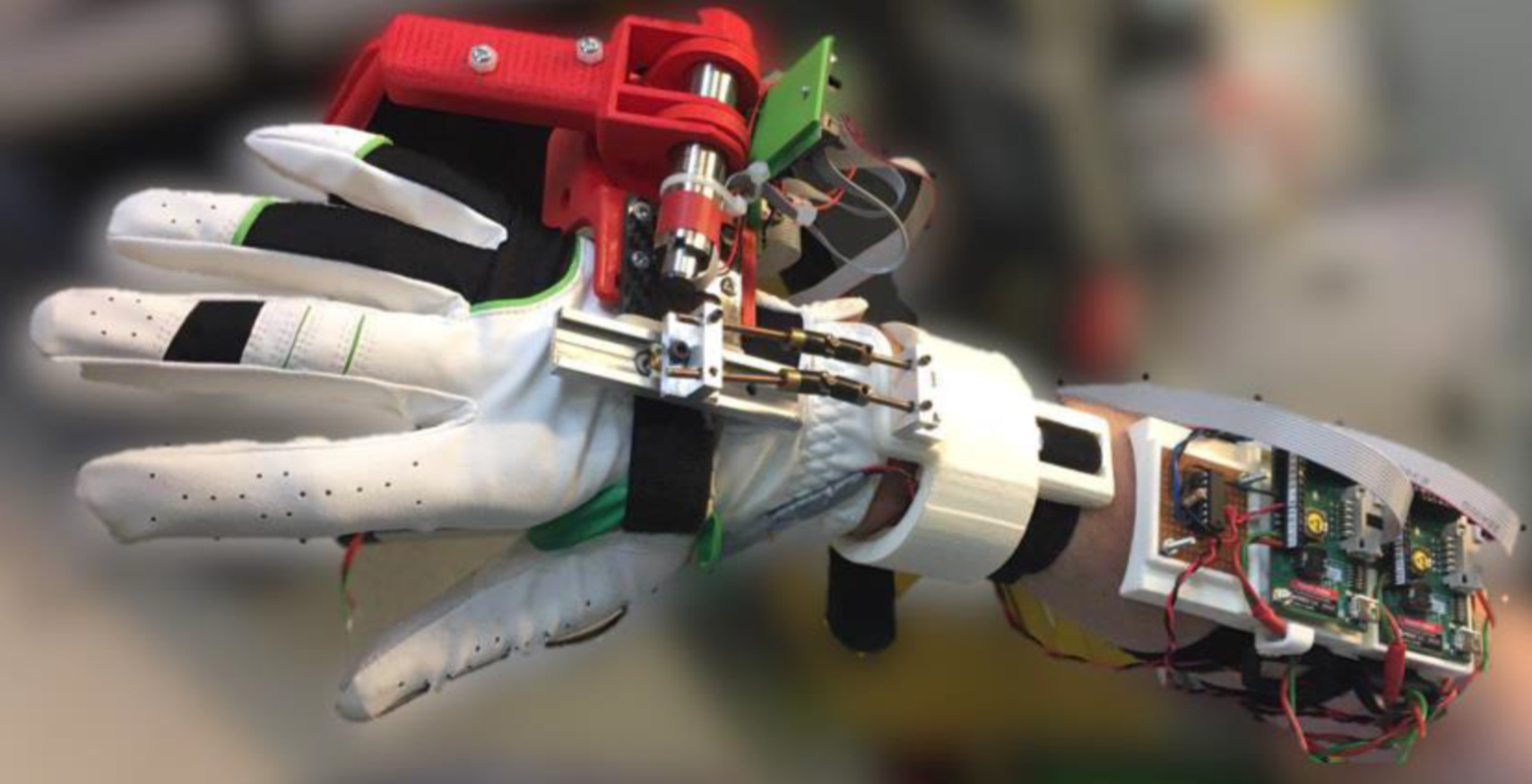}
\caption{Our supernumerary robotic 3\textsuperscript{rd} thumb (SR3T) in use.}
\label{intro}
\end{figure}
%%
%%%%%%%%%%%%%%%%%%%%%%%%%%%%%%%%%%%%%%%%%%%%%%%%%%%%%%%%%%%%%%%%%%%%%%%%%%%%%%%%
\vspace{-0.25cm}
\section{Background}

Wearable robotics can be seen as a very promising medium through which the human body and various robotic forms can be integrated for the purpose of synergistic performance of tasks. Wearable robotics come to place, for the most part, in the form of replacements for lost limbs, using prosthetics \cite{Zollo2007BiomechatronicApplications}, or as a means to enhance the human body's force generation mechanics and its ability to carry out precise motions using, for example, exoskeletons \cite{Zoss2006BiomechanicalBLEEX}.

A larger body of research exists related to supernumerary arms as they provide abilities that would be useful in a number of industrial settings. In \cite{Davenport2012DesignLimbs} two robotic arms that are mounted on a user's body, aiding them in performance of tasks. The SR arms have actuators which allow them to mimic the rotation and movement of the human arm. They are mounted to the waist of the user, so that minimum torque is exerted on the spine. The device can help in balancing or reduction of weight being handled by the user \cite{Parietti2014BracingReduction}. A demonstration control algorithm is employed by this design, a task is repeated a number of times by a worker, the identifier data of that task collected, and used to create a dynamic model for how the robotic “follower” arm should react to the user \cite{Llorens-Bonilla2012Demonstration-basedLimbs}. Because  the human body will not be normally stationary, be it through involuntary or voluntary movements, SR systems must be able to adapt in real-time. As an example, in \cite{Parietti2014BracingReduction}, a discrete Kalman filter is used to evaluate error covariance due to motion, and then use a bracing technique to counteract and suppress user induced disturbances. Bracing is carried out through attaching one of the two SR limbs to a structure in the environment, for support. In the setting of collaborating so closely with robots, the coordination between the human and the SR device is key. It has recently been shown that if algorithms can produce close mirroring of movements between humans and robots, the extra limbs can be perceived as part of the user's body \cite{Guterstam2011TheArm}. This would allow the user to interact with the SR peripheral more effectively and with less effort.

Research efforts have focused more on assistance with grasping or robotic limbs that help you while you work, with no strong body of work directed at creative applications of this technology, involving finer movements and force level control such as those with fingers. In \cite{Wu2014Bio-artificialFingers} a robotic aid in the grasping and handling of objects was developed using . They have proposed a system that synergizes grasp movements between human fingers and supernumerary robotics. This is done by completing grasps while wearing a high fidelity data glove which tracks wrist, fingertip and joint movements. In real-time use, a data glove is employed in order to relay hand movement information to the SR system so that it can react accordingly. Similar to this approach, \cite{Prattichizzo2014TheCapabilities} adds a single robotic finger to the user's hand that is positioned on the underside of the wrist forming a claw with the hand. In this research the robotic device reacts to movements of the user's hand which is mapped, again, using a data glove. The robotic finger, however, reacts in real-time to finger motions rather than preset configurations.

Research regarding supernumerary robotic fingers deals with the concept purely from a grasping perspective, and does not visit the need for applying forces, and variable force exertion in functional tasks. This is limiting when considering integration into uses other than picking up objects. Furthermore, the dexterity with which these systems can operate is limited and needs to be expanded to closer resemble human movement.
%%%%%%%%%%%%%%%%%%%%%%%%%%%%%%%%%%%%%%%%%%%%%%%%%%%%%%%%%%%%%%%%%%%%%%%%%%%%%%%%
\section{Methodology}
\begin{figure}[tp]
\centering
\includegraphics[width=0.8\columnwidth]{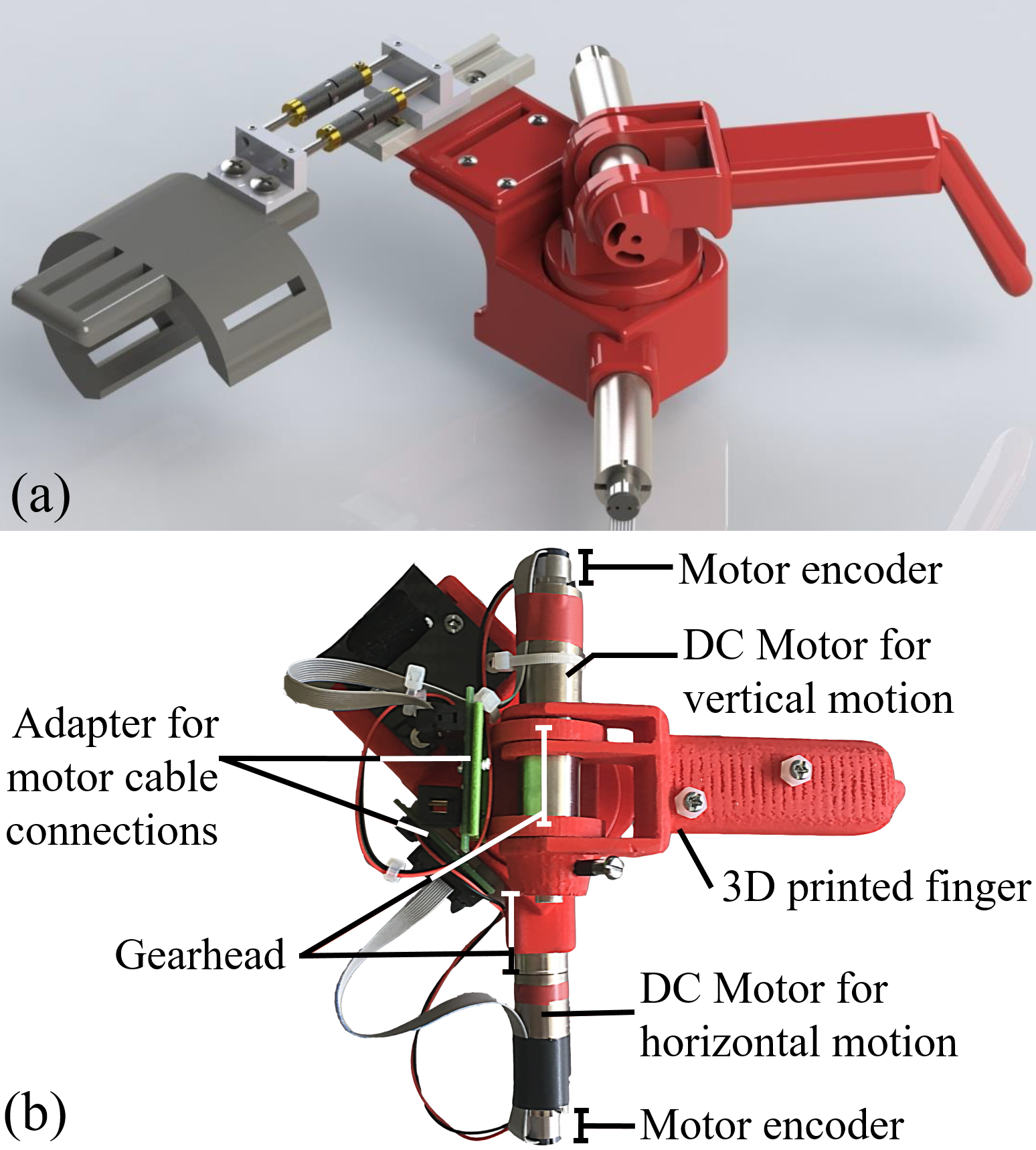}
\caption{The SR3T presented in this paper, (a) CAD drawing of the design in SolidWorks, (b) 3D printed and assembled device with labeling of the different parts.}
\label{design}
\end{figure}
We propose a supernumerary robotic 3\textsuperscript{rd} thumb (SR3T) which can augment the functionality of the human hand in dexterous tasks such as the playing of a piano, and to serve as a test rig for human augmentation through wearable robotics, allowing research into questions on embodiment and learn-ability of their use. A piano playing task is considered, as it allows for creative behavior from the user while using the SR3T in a typical functional human activity. The SR3T will thus have two degrees of freedom, allowing it to move horizontally (to reach a particular piano key) and vertically (to press the key). The user will control the two degrees of freedom of the SR3T using movements from other body parts, namely their thumb for horizontal and their foot for vertical alignment.

\subsection{Design specifications} To achieve the embodiment goals of the project, the following criteria are of high importance;

\underline{\smash{Human integration}}: The entire device must take a form and behave in a manner that would seem natural to the user as an addition to their existing fingers. This includes the physical appearance of the supernumerary finger, its control and its reaction to user input. Latency between user input and the movement of the finger/striking of a key is thus a critical metric to consider. Therefore, a total processing time threshold, from sensor input to motor actuation, is set at 80ms. This will allow a time window which will prevent a noticeable delay in hearing a sound/seeing movement \cite{Makin2017NeurocognitiveTechnology}.
\underline{\smash{Interference}}: As this supernumerary finger is to be used in the setting of playing a piano, it is key that the device does not interfere with the user's natural abilities. This includes the manner in which the device is mounted and the way it is controlled. The robotic finger should not cause limitations to the movements of the user's own fingers, with workspaces clearly separate from each other.
\underline{\smash{Device weight}}: The weight of the device is a very important metric in the situation where delicacy and agility are an essential part of the end-goal activity. Here, the user must be able to comfortably wear the device for a sustained period, suggested as between 45-60 minutes, continuously. Looking at pre-existing handheld devices for guidance on acceptable weight limits, it was found that modern Apple smartphones weigh between 148-202 grams (iPhone 8/Plus). As a device that users would bear the weight of for shorter periods of time than a phone, it was aimed to not exceed twice the average weight of a smartphone (~350g).
\underline{\smash{Accuracy}}: The SR3T must firstly be able to fit in the width of a standard piano key so that it can press a single key accurately, and secondly, must be able to move with a level of accuracy in rotation that adjacent black and white keys can be pressed in succession. The standard width of a white key on a piano is 23.5mm with a black key width of 13.7mm
%\footnote{\href{https://en.wikipedia.org/wiki/Musical_keyboard}{Wikipedia: Musical Keyboard}}
.
\underline{\smash{Torque}}: The key of a standard piano requires 0.5 N of force acting on its face to press it.
%\footnote{\href{http://www.pianofinders.com/educational/touchweight.htm}{http://www.pianofinders.com/educational/touchweight.htm}}. 
The motors used in the design should allow adequate torque to be generated such that this force requirement is met, and be adjustable as needed for different volume levels to be achieved.

\subsection{Mechanical implementation}

The SR3T consists of 2 degrees of freedom, with a finger-shaped body, moving vertically and horizontally as commanded by the user. Movements are performed by two 9V DC motors from Maxon Motors (Switzerland), customized with a planetary gearhead with reduction ratio 16:1, increasing the output torque 16 times higher than the nominal, by decreasing the velocity of the motor. The motors are equipped with the ENX10 EASY 256IMP encoders to read position and velocity. This is then used with the EPOS2 24/2 DC motor controllers, allowing control of motor current, position and velocity. These motor controllers also include analogue inputs, which are made use of for reading movement commands from the user. 

The main motivation when designing the shape of the finger attachment was that the user could naturally relate to its form. A large part of attempting to add to the basic human form is the user feeling that they may "embody" the robotic device that they are interacting with \cite{Makin2017NeurocognitiveTechnology}. Following this idea, different design ideas were considered regarding the size, shape and mechanical mechanisms that would be employed in creating the SR3T. From a functional stand point, the design must be able to carry out the task of pressing both white and black keys interchangeably, as the user sees fit. The finger design employed resembles that of a fixed human finger. It consists of a knuckle, around which rotation takes place, a horizontal section, similar to a human proximal phalange, and a hinged section, which represents a combined medial and distal phalange. The robotic finger was modeled on an average adult male's middle finger dimensions. The horizontal section has a length of 58mm, which joins the tip section of length 48.5mm, at an angle of 60 degrees. The knuckle section adds an additional 41mm of length to the finger, so that the positioning difference, back on the hypothenar eminence and the ulnar aspect of the hand, is offset. This allows the SR3T to be in line with the player's fingers. The player is then able to decide whether black or white keys are targeted through moving the whole hand in or out depth wise, on the piano keys. This is something that players augment inherently and so it is natural for them to implement in this scenario.

The design mounts flush to the hypothenar eminence, allowing the user to move their wrist freely and minimizing the distance between the device and the keys of the piano. The top of this mounting system is designed to sit over the back of the user's hand, with the large curvature on top and the smaller lip below following the curvature of a human hand for better fit and greater comfort. This in turn helps reduce any turning moments on the player's hand, generated by the weight of the device. The device sits across a 5cm long area, on the length of the hypothenar eminence. This dimension was used so that when the user attempts to press a piano key down, the finger mount is not pushed up by the resistance. For vertical rotation of the SR3T, a rotating top plate was designed. It was very important that the space taken up by this part, both horizontally and vertically, was minimized to achieve an overall design that kept the SR3T as close to the hand height as possible, while not adding a device with a cumbersome width. The rest of the device was recessed at different heights below this to ensure that the SR3T would eventually sit as close as possible to the user's fingers. The remainder of the design was driven by typical mechanical engineering criteria, ensuring proper mating of relevant parts and translating the rotations of the DC motors into the movements of the robotic finger. All necessary parts were designed in SolidWorks, 3D printed using PLA material, and assembled together.

The complete system can be seen in Figure \ref{design}. The horizontal motor is free to create a rotation of the finger at over 360 degrees, which is far beyond the needs of the piano player. The vertical motor is also free to create rotations up to about 120 degrees, allowing for the user to press piano keyboards with different levels of pressure with precision. The linear drive section of the wrist support contains two aluminum couplings, which were manufactured to specification using universal drives, and a plastic linear drive carriage and aluminum carriage rail to keep the different sections in place.

\subsection{Sensors and human interfacing}
\begin{figure}[tp]
\includegraphics[width=\columnwidth]{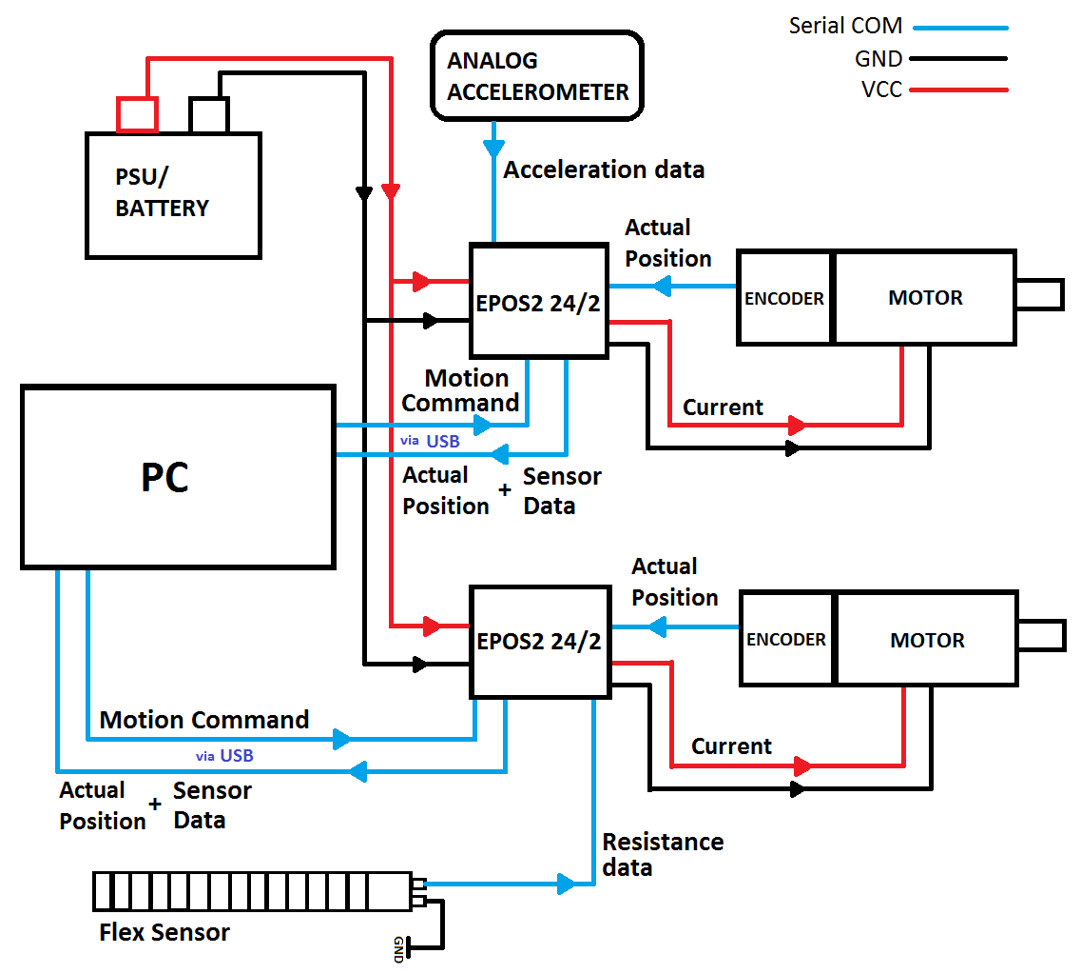}
\caption{Block diagram for the complete SR3T system.}
\label{blockdiagram}
\end{figure}
The SR3T's two degrees of freedom are controlled through two movements from the human user. The horizontal movement is picked up from the user's thumb on the hand being augmented. The vertical movement is picked up from the lifting movement of the user's left foot (dorsiflexion). The left foot is chosen due to the right foot being of use in pressing the pedals of the piano. 

To sense the flexion of the thumb and control the horizontal movement of the SR3T, a generic, off-the-shelf flex sensor is used. This is a variable resistor, which changes its resistance based on the curvature applied to it during flexion. When flat, the resistance value was measured at around 13 k\si{\ohm} and when at 180 degree pinch bending, the resistance value was measured at around 26 k\si{\ohm}. A voltage divider and unity buffer setup is used to interrogate the flex sensor, translating the changes in sensor resistance into buffer output voltage changes. This is then connected to the analog input of the EPOS2 24/2 motor controller for the horizontal movement. The flex sensor is placed on the thumb of a golf glove. The golf glove is made of leather material which provides flexibility and comfort to move the natural fingers. It also ensures that the flex sensor stays in place as the thumb moves, yet bends with respect to the motion of the thumb. 

To sense the dorsiflexion of the foot, the ADXL335 (Analog Devices, Massachusetts, USA) analog accelerometer is used. The device is mounted on the user's foot and its outputs connected directly to the analog inputs of the EPOS2 24/2 motor controller for the vertical movement. The values for two of the accelerometer axes are used. The Y axis provides information on the direction of the vertical movement and the Z axis provides information on the velocity of the movement. 

An arm mount for all the required printed circuit boards (PCBs) was designed and 3D printed, in order to make the system as portable as possible. The mount houses the two EPOS2 24/2 motor controllers and a smaller PCB for handling of sensor outputs; refer to Figure \ref{piano}.a.
\vspace{-0.25cm}
\subsection{Control software}

The control of the robotic system runs under a software algorithm that was built using C\# code in Visual Studio. The aim of the algorithm is to process data from sensors and convert them into robotic finger motion command. In order to speed up the computation time, the vertical motion algorithm and the horizontal motion algorithm run on separate threads in the C\# code. This method allows the calculation for vertical motions and the horizontal motions to run in parallel. The sensors are connected to the motor controllers, the PC reads sensor data via the analog inputs of the motor controllers, computes all input information from sensors and sends control commands back to the motor controllers. The motor controllers are connected to the PC via USB, refer to Figure \ref{blockdiagram}. The software presents a graphical user interface (GUI) for the user to configure and calibrate the SR3T as needed.

When the motor controller is enabled, the position value from the encoder will be restarted to zero. Consequently, the position of the SR3T at the beginning of the experiment determines the start of the position counter of the encoder. Using this method, the motion control calculations remain unaffected by the starting position of the device. In the horizontal movement calibration, the furthest piano note from the user's hand that the SR3T can reach will be selected as the maximum, and the closest piano note to the user's hand that the SR3T may press, as the minimum. In the horizontal motion calibration, the analog flex sensor values that correspond to the maximum and minimum flexion of the user's thumb are also recorded. The vertical and horizontal motion parameters are saved by moving the robotic finger to the preferred positions and triggering a button on the software GUI. The analog maximum and minimum parameters for the flex sensor are saved by straightening and flexing the thumb to the preferred posture. The velocity of the horizontal movement is defined through a proportional control approach, i.e. a larger distance from the SR3T's initial position to its target position, will result in a larger velocity of motion.

% %%
% \begin{figure}[tp]
% \includegraphics[width=\columnwidth]{Fig4.png}
% \caption{The user controlling the SR3T through sensors. Vertical movement: (a) SR3T lifted by lifting the foot, (b) SR3T lowered by putting the foot down. Horizontal movement: (c) SR3T extended outwards by extending the thumb outwards, (d) SR3T brought inwards by bending the thumb inwards.}
% \label{control}
% \end{figure}
% %%
In the vertical motion calibration, the highest vertical position and the lowest vertical position are set as maximum and minimum respectively. The highest vertical position is chosen as the position when the robotic finger hovers above the piano notes while the lowest vertical position is chosen at the height at which the SR3T has fully pressed a piano key. Y-axis acceleration data determines the position of the foot above the ground. This data corresponds to the direction of the SR3T's vertical motion. Z-axis acceleration data describes the instantaneous acceleration of the foot when it starts moving. This data is related to the velocity of the vertical motion of the SR3T. The Y-axis maximum and minimum parameters are saved by moving the foot up and by placing the foot on the ground respectively. Z-axis maximum parameter is the acceleration data when the foot is being lifted and Z-axis minimum parameter is the acceleration data when the foot is stationary.

The calibration processes above maps the sensor input and motor outputs in a linear fashion, allowing the user to control the position and velocity of the SR3T's movement, in the two defined degrees of freedom.

\section{Experiments and Discussion}
\subsection{Robot design} 
\begin{figure}[tp]
\includegraphics[width=\columnwidth]{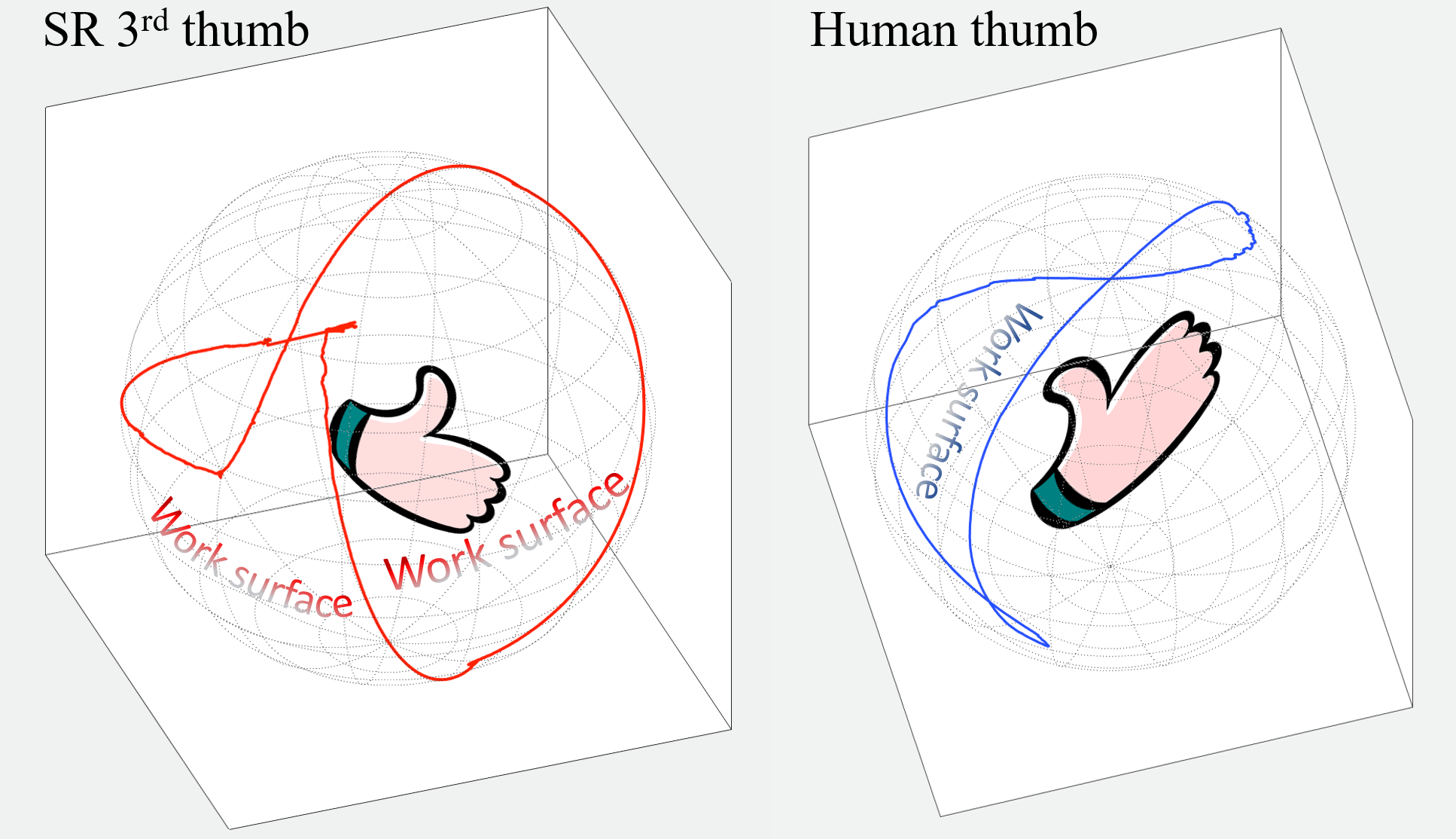}
\caption{Angular end-point workspace of the SR3T compared to that of a human thumb, based on data collected through optical motion trackers, and mapped onto unit sphere.}
\label{thumbdata}
\end{figure}

Experiments were performed to evlauate the SR3T's workspace. An optical motion capture system, the OptiTrack Prime 13W (NaturalPoint, Inc. DBA OptiTrack, Oregon, USA), was used to capture the 3D motion of the thumb and the SR3T in real time. The cameras captured the change in body position provided by the specific optical markers attached. During the experiment, three OptiTrack cameras were used to track the motion of the thumb in $x$, $y$, and $z$ planes. Before the motion tracking experiment started, the OptiTrack cameras were calibrated to ensure the accuracy and precision of the recorded data positions. Markers were placed at the base and tip of the SR3T, to measure its maximum workspace, sweeping the SR3T along its maximum range of movement. The same experiment repeated, with the subject moving their thumb along its maximum range of movement, and its motion tracked. Results are visualised in Figure \ref{thumbdata}, with the angular end-point surface mapped onto an sphere. Overall, the SR3T has an angular end-point surface 4 times that of the human thumb, effectively augmenting human reach and capability.

Using the same setup, delay experiments were also performed, to measure the delay between the command given by the user and the respective motion by the SR3T, i.e. thumb extension/flexion for horizontal and foot lift and drop for the vertical motions. Optical markers were placed on the SR3T, the user's thumb and their foot. Results showed a mean delay between motor intention and motor action of $\approx 85$msec.

\subsection{Human augmentation}
Our proof-of-concept test is designed to evaluate a naive users performance when using the SR3T in a functional manner and to explore whether the SR3T is used so as to substitute the thumb on the ipsilateral hand (i.e. using the right hand as if it were a left hand) or whether the user can embrace the ability to operate a hand with six fingers. To this end we focused on the task of piano playing which allows straightforward dynamic mapping and remapping of fingers to skilled task performance. In music, the choice of which fingers and hand positions to use when playing certain musical instruments is called fingering. Fingering typically is dynamically adjusted throughout a musical piece by skilled performers so as so optimally use the available fingers. The challenge of choosing a good fingering is to make the hand movements as comfortable as possible in the flow of the music without changing hand position too often. 

The test aims to evaluate whether an end-user would adapt and learn to use the device as an extension to their own fingers. A musician who had no previous interaction with the SR3T was asked to play piano with the device mounted on their forearm. The system control was explained to them and, over a period of two hours, the musician was allowed to improvise and try to use the finger in different musical contexts. At the beginning of the test, the user was struggling to control the SR3T to press even a single note, while doing nothing else with either hand, due to having to focus on moving their thumb and foot in correct coordination to control the SR3T. Over the course of the two hours, the musician went from this stage, to being able to use the SR3T to play melodies and chords, while using all of their other natural fingers simultaneously. The pianist demonstrated the ability to use the added finger, and the extra reach it provided, in a musical fashion, adding higher notes into the chords of the left hand and playing additional notes with melodies. 

In the first few trials it was observed that the SR3T presses the piano notes at considerable delay. But as the experiment went on, the delay decreased. This result suggests that the delays occurred as the pianist hesitated, focusing and move their foot to press the piano note using the SR3T. After a while, the cognitive load decreases and the pianist is able to move their thumb and foot naturally to control the robotic finger in real time. The results of the experiment confirmed our proposal that using the SR3T, a pianist is able to reach piano notes beyond their natural hand-span. The SR3T was able to press as far as 4 whole notes beyond the hand-span of the pianist. The results also showed that the pianist was able to play improvised musical pieces with 11 fingers (10 natural fingers and 1 robotic finger). The pianist was also able to press the black piano key notes using the SR3T, although it was more difficult to press those compared to the white piano note keys.
\begin{figure}[bp]
\centering
\includegraphics[width=0.8\columnwidth]{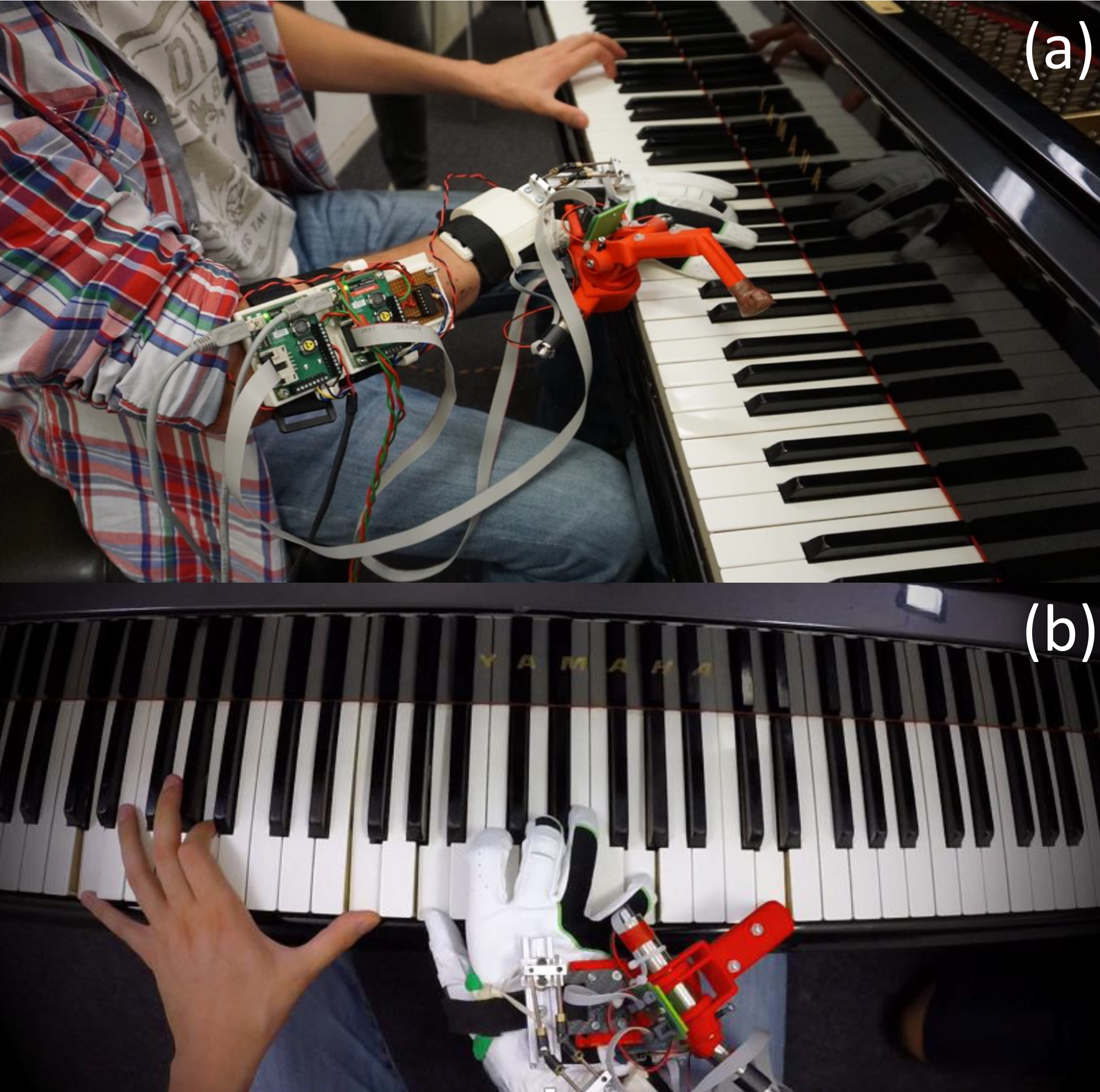}
\caption{The test subject playing piano using the SR3T. (a) view of the full system mounted on the user's arm as they play piano, (b) user's point of view; note the 4-note increase in reach using the SR3T.}
\label{piano}
\end{figure}
\noindent{\underline{\smash{Range Increase}}: The main aim was to increase the player's range. During the course of the test it was found that the SR3T could play notes, from a single semitone to three and a half semitones to the right of the player's pinkie finger. This is in effect a four whole-note increase in range to the musician's natural hand span. This can be seen in Figure \ref{piano}.b, where the player uses their thumb to play, whilst controlling the SR3T. Note that the SR3T extends the reach while still using the 5 fingers of the hand, adding an extra note to the reach of the user. If the hand is fully extended already, however, the extent of note reach augmentation will be reduced, due to the mounting style.
\noindent\underline{\smash{Dynamics}}: The pianist was able to control the dynamics of how hard or soft they press piano keys with the SR3T. This was achieved by lifting their foot, which had the accelerometer attached to it, at different speeds, which were reflected on the SR3T's movement.
\noindent\underline{\smash{Accuracy}}: The musician was able to play both black and white keys accurately using the finger. The finger was also capable of moving between adjacent keys in succession. The user pressing a black key can be seen in Figure \ref{piano}.a.
\noindent\underline{\smash{Comfort}}: The test went on for over two hours, with the pianist wearing the finger and attachments at all times. When asked during, and after the test, the user reported that they suffered little to no discomfort throughout.
\noindent\underline{\smash{Delay}}: The user was able to play in time with the additional finger, after the earlier stage of practice and learning. The initial delay is attributable to the user learning to rely on the movements of their foot to play a key, rather than due to delays in the robotic system.
\noindent\underline{\smash{Interference}}: Although the user initially had to think a lot about how to position their hand, the device did not interfere with their natural ability to play the instrument after the initial adjustment to its presence and was used seamlessly afterwards.

% To investigate the behaviour of the flex sensor, the markers were attached at each joint of the thumb to obtain the position of thumb joints for each time steps. A set of 3 additional markers were placed on the back of the hand to be detected as a rigid body and a reference point. The voltage data from the flex sensor was also recorded using an NI-USB 6300 and MATLAB code. The total duration to run this experiment was 2 minutes, with specific thumb flexion periods of 2 seconds each.
\vspace{-0.25cm}
\section{Discussion \& Conclusions}
\vspace{-0.1cm}
This paper presents the development of a supernumerary robotic 3\textsuperscript{rd} thumb for the purpose of functional human augmentation in a skilled motor task: piano playing. The mechanical structure, electronic components and control strategy to be used in the execution of the SR3T were described in detail. The SR3T's angular end-point surface was 4 times that of a human thumb, and mean motor intention to action delay was measured as $\approx 85$msec. A functional user test with a naive pianist showed that the SR3T was in fact effective, and that the user could adapt to incorporate it into their natural playing of the musical instrument. The piano player transitioned from an initial use of the SR3T on the right hand as if it were a left hand, using 5-finger based fingering, and managed after 1 hour to use 6-finger based fingering, i.e. using the right hand as a truly augmented hand.  

Given the results of our preliminary functionality test we conclude that there is a positive outlook for future work. The user demonstrated the ability to learn and control the SR3T quickly, and was able to produce musical results in hours which is little time compared to rehabilitation times for bionic prosthetics users of months. Furthermore, exploring the idea of low-dimensional control synergies , such as those employed in \cite{Wu2014Bio-artificialFingers,konnaris2016sparse}, might result in a better performance of the robotic finger in action.
While our human-robot interface is substitutive in nature (using other limb movements to control artificial movements\cite{thomik2013real,belic2015decoding}), true augmentation may require more sophisticated human-robot interfaces that are not just neurobionic \cite{Fara2014PredictionMMG} but cognitive in nature \cite{maimon2017towards}.

The SR3T setup provides a unique test rig for experiments on learn-ability and embodiment potential of supernumerary robotic limbs and augmentative devices. The piano application presents a creative environment where the user is free to improvise and learn along the way how to (or not to) incorporate the SR3T into their work, while focusing on the control of two extra degrees of freedom.Future work will focus on improving the setup and running extended experiments to better understand the implications and limitations of robotic augmentation. 
Our work here can be considered  an "existence proof" that naive, but hand-craft skilled users can indeed learn in under a few hours to perform highly dexterous tasks such as piano playing, suggesting that our brain has sufficient plasticity and capacity to support true human augmentation through supernumerary robotic limbs. 

%%%%%%%%%%%%%%%%%%%%%%%%%%%%%%%%%%%%%%%%%%%%%%%%%%%%%%%%%%%%%%%%%%%%%%%%%%%%%%%%
\vspace{-0.15cm}

\bibliographystyle{ieeetr}
\bibliography{Mendeley.bib,additional.bib}

\end{document}